\documentclass{article}
\PassOptionsToPackage{numbers, compress}{natbib}

    \usepackage[final]{neurips_2023}

\usepackage[utf8]{inputenc} % allow utf-8 input
\usepackage[T1]{fontenc}    % use 8-bit T1 fonts
\usepackage{hyperref}       % hyperlinks
\usepackage{url}            % simple URL typesetting
\usepackage{booktabs}       % professional-quality tables
\usepackage{amsfonts}       % blackboard math symbols
\usepackage{nicefrac}       % compact symbols for 1/2, etc.
\usepackage{microtype}      % microtypography
\usepackage{xcolor}         % colors

\usepackage{algorithm}
\usepackage{algpseudocode}

\usepackage{caption}
\usepackage{subcaption}

\usepackage{amsmath}
\usepackage{amssymb}
\usepackage{mathtools}
\usepackage{amsthm}

\theoremstyle{plain}
\newtheorem{theorem}{Theorem}[section]
\newtheorem{proposition}[theorem]{Proposition}
\newtheorem{conjecture}[theorem]{Conjecture}

\theoremstyle{definition}
\newtheorem{definition}[theorem]{Definition}
\newtheorem{assumption}[theorem]{Assumption}
\theoremstyle{remark}

\DeclareMathOperator*{\argmin}{arg\,min}
\DeclareMathOperator{\Tr}{\text{Tr}}

\newcommand{\shorteq}{%
  \settowidth{\@tempdima}{-}% Width of hyphen
  \resizebox{\@tempdima}{\height}{=}%
}

\title{Adversarial Estimation of Topological Dimension\\with Harmonic Score Maps}

\author{%
  Eric Yeats\thanks{Work conducted while at Oak Ridge National Laboratory through a DOE-SCGSR Award}\\
  Duke University\\
  \texttt{eric.yeats@duke.edu} \\
  \And
  Cameron Darwin \\
  \texttt{cmdarwin@gmail.com} \\
  \AND
  Frank Liu \\
  Old Dominion University  \\
  \texttt{fliu@odu.edu} \\
  \And
  Hai Li \\
  Duke University \\
  \texttt{hai.li@duke.edu} \\
}

\begin{document}

\maketitle

\begin{abstract}
  Quantification of the number of variables needed to locally explain complex data is often the first step to better understanding it. Existing techniques from intrinsic dimension estimation leverage statistical models to glean this information from samples within a neighborhood. However, existing methods often rely on well-picked hyperparameters and ample data as manifold dimension and curvature increases. Leveraging insight into the fixed point of the score matching objective as the score map is regularized by its Dirichlet energy, we show that it is possible to retrieve the topological dimension of the manifold learned by the score map. We then introduce a novel method to measure the learned manifold's topological dimension (i.e., local intrinsic dimension) using adversarial attacks, thereby generating useful interpretations of the learned manifold.
\end{abstract}

\section{Introduction}

Topological dimension (TD) estimation, or local intrinsic dimension estimation, computes the number of dimensions of variation of data near a point. The topological dimension plays a central role in data compression and data understanding, and it is linked to the generalization ability of classifiers in machine learning \cite{campadelli2015intrinsic, pope2021intrinsic}. Due to its importance, many methods have been developed to estimate topological dimension based on lower-dimensional projection, fractal dimensions, or statistical fitting \cite{rozza2012novel, levina2004maximum}. Common weaknesses of existing methods are that they require carefully selected hyperparameters, have limited scalability to high dimensions, and lack robustness to noise \cite{camastra2016intrinsic}.

Score based models (SBMs) \cite{song2019generative, song2020score} form the leading generative method for continuous \cite{ho2020denoising} and graph \cite{vignac2022digress} data. A popular method to train score based models is by denoising score matching \cite{vincent2011connection}, which scales well to high-dimensional, complex distributions. However, like other deep learning approaches, SBM performance (in tems of likelihood) tends to be sensitive to malicious input perturbations \cite{szegedy2013intriguing, yeats2021improving} and SBMs require interventions to be more interpretable \cite{lee2023diffusion, yeats2022nashae}.

In light of these limitations, we identify the \textit{locally averaging} property as key to instilling robustness in learned score maps, and we augment the denoising score matching objective with Dirichlet energy regularization to realize it. Under reasonable assumptions, we show that Dirichlet energy regularization of SBMs simultaneously boosts adversarial robustness while revealing the underlying topological dimension of the learned manifold. We prove that Dirichlet energy regularization corresponds precisely to learning additional variance normal to the manifold, leading to a novel method to reveal the SBM's learned topological dimension using adversarial attacks.

\section{Methodology}

\paragraph{Local Averaging Property} Log-likelihood adversaries occur where a small perturbation results in a large change in log-likelihood which does not reflect the true distribution \cite{pope2020adversarial}. The potential to have a log-likelihood adversary, then, is controlled by the score. 

To enhance the robustness of learned score maps, we suggest to promote the locally-averaging property. That is, to enforce an additional constraint on the score map which requires the score at any given point to be close to the average score in an $\varepsilon$-neighborhood of that point. See appendix (\ref{app:local_averaging}) for our argument.
\begin{conjecture}
\textbf{(Local Averaging)} Consider a score objective $S$ used to evaluate the performance of a model $M$ given a data sample $\mathcal{D}$. I.e., a lower value of $S(M,\mathcal{D})$ means better estimated performance. Let $LV(M)$ be a functional measuring how far a score map is from being locally averaging. E.g., this could be the uniform norm
\[
\sup_x \left|M(x) - \mathbb{E}_{|x' - x| < \varepsilon} M(x')\right|.
\]
Then for another score model $M'$, with $S(M',\mathcal{D}) = S(M,\mathcal{D})$, if $LV(M') < LV(M)$, then $M'$ will possess fewer log-likelihood adversaries than $M$, and be more robust to changes in the original sample. Moreover, if $M$ is modified to reduce $LV(M)$, then $M$ will be changed more in areas of low density than in areas of high density, so that the reduction in log-likelihood adversaries will be proportionally greater than the reduction in the performance of $M$ according to $S$.
\end{conjecture}
It is well-known that harmonic functions have the \textit{averaging} property such that at any point, the value of the function at the point is equal to the average of the function within a ball centered on the point. A map $\phi: X \rightarrow Y$ is harmonic if it is a critical point of the Dirichlet energy (in the space of maps which satisfy specified constraints on the boundary).
\begin{definition}
\label{def:dirichlet_energy}
\textbf{(Dirichlet Energy)} The Dirichlet energy (DE) of a map $\phi: X \rightarrow Y$ is:
\[
    \mathcal{D}[\phi] = \frac{1}{2} \int_{X} \|d\phi_x\|^2 d\text{Vol}(x),
\]
where $\|d\phi_x\|$ is the spectral norm of the differential of the map at $x$.
\end{definition}
Framing this in the context of SBMs, the score matching objective can be thought of as a soft boundary constraint for each score map in time. As the Dirichlet energy of each score map is reduced subject to the soft boundary constraints provided by the training data, the score map is encouraged to become a \textit{minimal surface} \cite{dherin2022neural}. % Much like how minimizing the path length between two points yields a line, minimizing the surface area of a score map should prevent its learned score from exploding when it is not called for by the training data, thereby mitigating adversarial exploits for a SBM.
\begin{proposition}\label{prop:gaussian_dirichlet_energy}
    \textbf{(Gaussian Score DE)} Reducing the Dirichlet energy of a multivariate Gaussian score map is to increase its variance (resp. entropy) along the direction(s) of minimal variance.
\end{proposition}
\begin{assumption}\label{ass:gauss}
    \textbf{(Smooth Manifold with Gaussian Noise)} The ambient space around a point $x$ near the manifold can be \textit{locally} characterized by two orthogonal subspaces: the tangent subspace $X_\mathcal{M}$ and the normal subspace $X_\perp$ ($X:=X_\mathcal{M}\oplus X_\perp$). Hence $x=(x_\mathcal{M}, x_\perp)$, where $x_\mathcal{M}$ and $x_\perp$ are subsets of $x$ expressed in the bases of the manifold tangent and normal subspaces. Furthermore, the local density around the manifold can be expressed as $p(x) = p_{X_\perp}(x_\perp)p_{X_\mathcal{M}}(x_\mathcal{M})$, where $p_{X_\perp}(x_\perp)$ is isotropic Gaussian with covariance $\sigma^2 I$ and $\mu=0$.
\end{assumption}
\begin{theorem}\label{thm:additive_variance}
    \textbf{(Additive Variance Property)} In an ambient space of dimension $n$, suppose a score model is regularized with $n\gamma$ level of DE on training data near the manifold. Furthermore, the local normal subspace has dimension $n_\perp$ (where $0 < n_\perp \leq n$) and the distribution which occupies it is isotropic Gaussian with variance $\Sigma=\sigma^2I$ (Assumption \ref{ass:gauss}). In the asymptotic setting, the regularized score map learns an ambient density with local off-manifold variance $(\sigma^2 + (n / n_\perp) \gamma)I$.
\end{theorem}
\paragraph{Method to Calculate Topological Dimension} Theorem (\ref{thm:additive_variance}) relates the level of Dirichlet energy regularization to the normal subspace variance. Its proof is in appendix (\ref{app:additive_variance}). Therefore, if one has knowledge of the values of $\gamma$, $n$, and the noise variance $\sigma$, one can employ adversarial attacks to measure the learned off-manifold variance and thereby recover the topological dimension. \textbf{Algorithm (\ref{alg:topological_dimension_est})} depicts the topological dimension estimate for an $L2$-bounded adversarial function $\Tilde{g}(x, \epsilon)$.
\begin{algorithm}
\caption{Topological Dimension Estimate}\label{alg:topological_dimension_est}
\begin{algorithmic}
\Require $x \in \mathbb{R}^n,\ s_{\theta}: \mathbb{R}^n \rightarrow \mathbb{R}^n,\ \Tilde{g}:\mathbb{R}^n \times \mathbb{R}^+ \rightarrow \mathbb{R}^n, \ \gamma \in \mathbb{R}^+,\ \sigma \in \mathbb{R}^+$ \\ \Comment{$s_{\theta}(x)$: score map with $\gamma$ DE reg. and $\sigma$ noise scale, $\Tilde{g}(x, \epsilon)$: attack func. with $L2$ budget $\epsilon$}
\Ensure $\hat{n}_\mathcal{M} \approx n_\mathcal{M}$
\State $\Tilde{x} \gets \Tilde{g}(x, \sigma)$ \Comment{$\Tilde{g}$ returns an adversarial example near $x$ (e.g., by following $-s_\theta(x)$)}
\State $\delta \gets \left\|s_\theta(\Tilde{x}) - s_\theta(x)\right\| / \left\|\Tilde{x} - x\right\|$ \Comment{$\delta$ stores learned score ``slope''}
\State $\hat{n}_\mathcal{M} \gets n - n\gamma/(\delta^{-1} - \sigma^2)$ \Comment{$\delta$ should be approximately $1/(\sigma^2 + n\gamma/n_\perp$)} 
\State \textbf{return} $\hat{n}_\mathcal{M}$
\end{algorithmic}
\end{algorithm}
\paragraph{Training Implementation} We augment weighted denoising score matching with DE regularization:
\[
    \theta^* = \argmin_\theta \mathbb{E}_t \lambda(t) \mathbb{E}_{x_0}\mathbb{E}_{x_t|x_0} \, \| s_\theta(x_t, t) - \nabla_{x_t} \log p_{0t}(x_t | x_0) \|^2 + n\gamma \|ds_\theta(x_t, t)\|^2
\]
Here, $\|ds_\theta(x_t, t)\|$ is the spectral norm of the score differential at $(x_t, t)$, $n$ is the ambient dimension of $X$, and $\gamma$ controls the strength of Dirichlet energy regularization. For SBM experiments, we use the continuous variance preserving denoising diffusion probabilistic model (DDPM) of \citet{song2020score}. Experiments on individual score maps employ a single noise scale $\sigma$ and have no time dependence. The spectral norm is implemented efficiently by our Jacobian power iteration (see appendix \ref{app:jac_pow_it}).
\section{Experiments}
\begin{figure*}[h]
    \centering
    \begin{subfigure}[t]{0.49\textwidth}
        \includegraphics[width=\columnwidth]{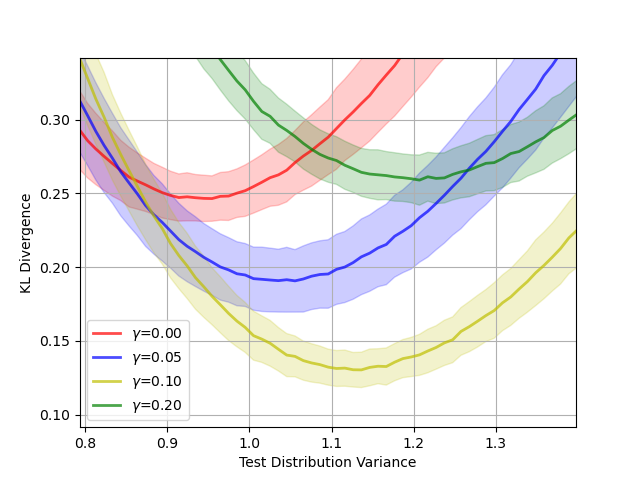}
        \caption{KL divergence of isotropic Gaussian distributions of different variance with the learned distributions of DDPMs. Solid line: average, shaded: one stdev.}\label{fig:add_var_test}
    \end{subfigure}
    \hfill
    \begin{subfigure}[t]{0.49\textwidth}
        \includegraphics[width=\columnwidth]{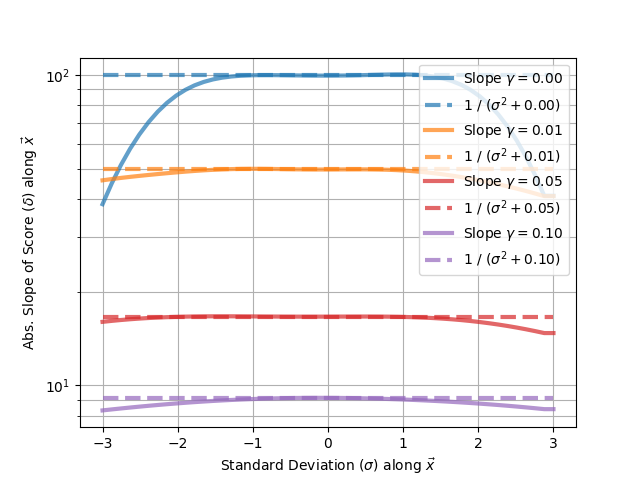}
        \caption{Comparison of learned score ``slope'' along the vector $\Vec{x}=\Vec{1}$ of score maps trained on an isolated point in $\mathbb{R}^{16}$ with $\sigma=0.1$ and various levels of DE reg. ($\gamma$).}\label{fig:add_var_map}
    \end{subfigure}
    \caption{Experiments measuring the additive variance property.}
\end{figure*}

\begin{figure*}
    \centering
    \hfill
    \begin{subfigure}[t]{0.4\textwidth}
        \includegraphics[width=\columnwidth]
        {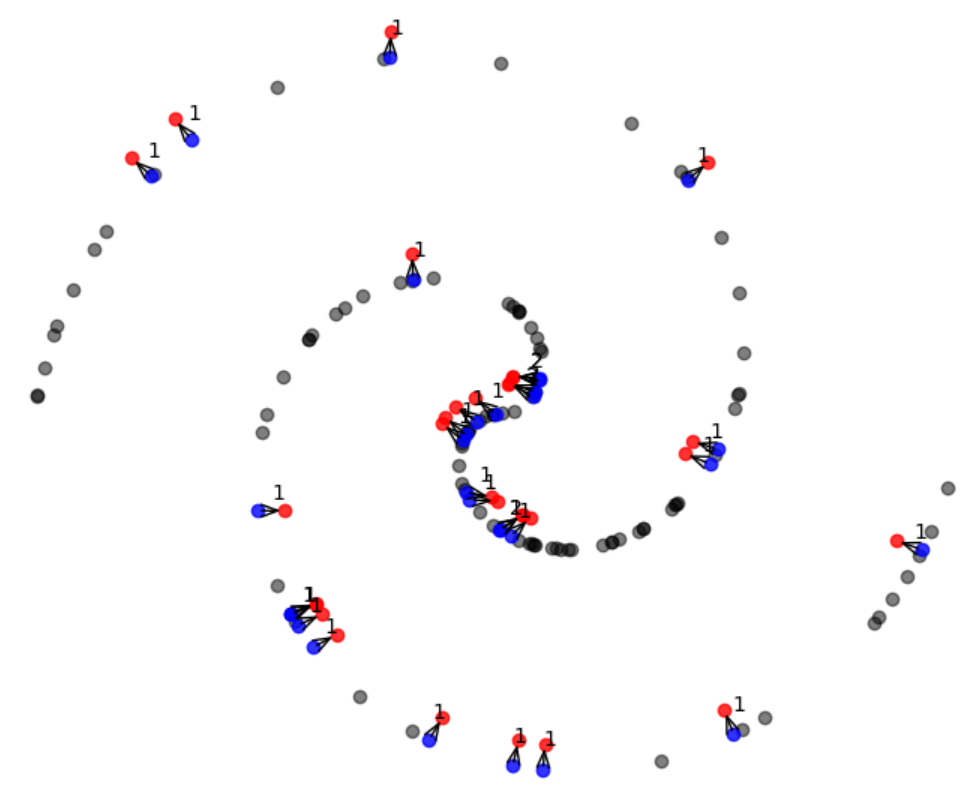}
        \caption{Depiction of our method on ``Swirl'' data. Some randomly selected \textcolor{blue}{original data} and \textcolor{red}{adversarial data} are plotted with associated estimates of the topological dimension.
        }\label{fig:swirl_attack}
    \end{subfigure}
    \hfill
    \begin{subfigure}[t]{0.45\textwidth}
        \includegraphics[width=\textwidth]{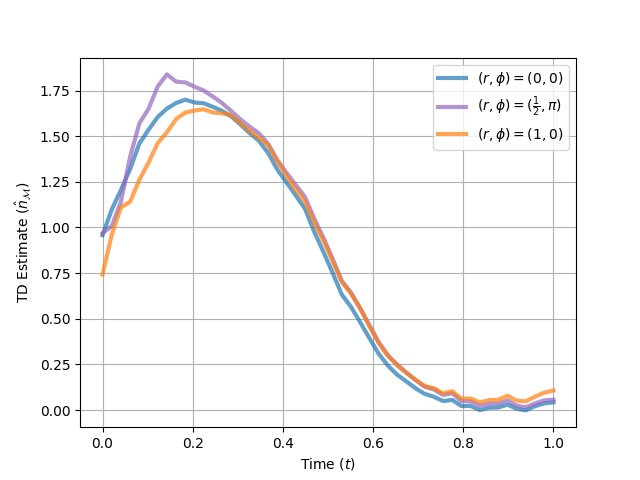}
        \caption{TD estimates over time given by DDPM score maps for selected ``Swirl'' data points. Key: $(r,\phi)$ are polar coordinates relative to the swirl center, with $r=1$ corresponding to distant points.}\label{fig:lid_time}
    \end{subfigure}
    \hfill
    \caption{Our topological dimension estimation method applied to the ``Swirl'' manifold.}\label{fig:ldb}
\end{figure*}

\paragraph{Adversarial Attacks on CIFAR-10} Dirichlet energy regularization allows us to control the off-manifold (normal) variance of a learned distribution and thereby mitigate log-likelihood adversaries. Table (\ref{tab:adver_cifar}) depicts results of PGD adversarial attacks \cite{madry2017towards} on DDPMs trained on CIFAR-10. Dirichlet energy regularized diffusion models are associated with increased BPD on the standard (``clean'') test dataset, but they exhibit improved robustness to perturbed data. Compared with similar methods such as normalizing flows \cite{pope2020adversarial} on CIFAR-10, DDPMs seem inherently more robust. \vspace{-10pt}
\begin{table}[h]
  \caption{BPD($\downarrow$) for $L_2$ attacks on CIFAR-10 test set. \textbf{Key:} Method(iters, $\epsilon$)}
  \label{tab:adver_cifar}
  \centering
  \begin{tabular}{lccccc}
    \toprule
    DE Reg. ($\gamma$) & Clean & Random(1, 0.2) & PGD(1, 0.2) & Random(1, 0.8) & PGD(20, 0.8) \\
    \midrule
    0 & \textbf{3.288} & 3.81 & 4.182 & 4.834 & 5.282 \\
    1e-4 & 3.387 & \textbf{3.776} & 4.127 & 4.803 & 5.211  \\
    2e-4 & 3.464 & 3.793 & \textbf{4.118} & \textbf{4.787} & \textbf{5.187} \\
    \bottomrule
  \end{tabular}
\end{table}\vspace{-10pt}
\paragraph{Additive Variance Property} DDPMs are trained on the same $8$-dimensional Gaussian distribution $x_1 \sim \text{Gauss}(X_1;\, 0, I)$ with varying levels of DE regularization. Then, their KL divergence with Gaussians of different variance $x_{\sigma} \sim \text{Gauss}(X_{\sigma};\, 0, \sigma^2I)$ is measured and reported in \textbf{Figure (\ref{fig:add_var_test})} for 20 trials. There is a clear one-to-one relationship between the level of DE regularization and the additional variance learned by the DDPM: the learned distributions have minimal KL-divergence with test distributions of variance $\sigma^2 \approx 1+\gamma$ although their training data had unit variance. In \textbf{Figure (\ref{fig:add_var_map})}, individual score maps are trained on an isolated point in $\mathbb{R}^{16}$ with $\sigma=0.1$ and differing $\gamma$. The ``slope'' of each score map along the vector $\Vec{x}=\Vec{1}$ indicates that the learned variance is $\sigma^2 + \gamma$. Here, the ``slope'' is $\delta$ from \textbf{Algorithm (\ref{alg:topological_dimension_est})}, but along $\Vec{x}$ rather than an adversarial sample. \vspace{-10pt}
\paragraph{Estimation of Topological Dimension}
\textbf{Figure (\ref{fig:swirl_attack})} visualizes the adversarial TD estimation method for some samples from the ``Swirl'' dataset. The method returns accurate estimates despite the curvature and non-uniform sampling of the manifold. \textbf{Figure (\ref{fig:lid_time})} depicts TD estimates of selected ``Swirl'' points as they are decayed (diffused via VP SDE without noise) through time. The TD estimates reflect the compression of the locally 1d swirl structure to a disk and then eventually to a point (the Gaussian prior). \textbf{Table (\ref{tab:topological_dim})} reports mean squared error (MSE) results of TD estimates for the ``Swirl'', ``LineDiskBall'', and ``HyperTwinPeaks'' manifold datasets. We incorporate results from two $k$-nearest neighbor statistical methods: MLE \cite{levina2004maximum} and $\text{MiND}_\text{ML}$ \cite{rozza2012novel} from scikit-dimension \cite{bac2021scikit}. In the table, subscripts of MLE and MiND indicate the number of $k$-nearest neighbors used. This hyperparameter is crucial to ensure good performance of the statistical methods. $\text{SM}_{0.01}$ indicates that the score map was trained with $\gamma=0.01$ DE regularization ($\sigma=0.1$). For the ``Swirl'' dataset, we report the results without noise and with added Gaussian noise of scale $0.01$. Our method is competitive with the statistical approaches on the simpler ``Swirl'' and ``LineDiskBall'' manifolds, and it is more accurate on noisy ``Swirl'' and the higher-dimensional ``HyperTwinPeaks'' manifolds.
\begin{table}[h!]
  \caption{MSE ($\downarrow$) of topological dimension prediction averaged over 5 independent trials}
  \label{tab:topological_dim}
  \centering
  \begin{tabular}{lcccccc}
    \toprule
    Benchmark & TD(s) & $\text{MLE}_{10}$ & $\text{MLE}_{20}$ & $\text{MiND}_{10}$ & $\text{MiND}_{20}$ & $\text{SM}_{0.01}$ \\ % DDPM_{0.01}
    \midrule
    $\text{Swirl}$ & 1 & 0.152 & 0.063 & 0.171 & \textbf{0.001} & 0.046 \\ % 0.309
    $\text{Swirl}\ \sigma_{0.01}$ & 1 & 0.495 & 0.272 & 0.974 & 0.486 & \textbf{0.080} \\ % 0.364
    LineDiskBall & 1-3 & 0.308 & 0.226 & \textbf{0.118} & 0.479 & 0.312 \\ % 0.251
    HyperTwinPeaks & 10 & 5.931 & 5.006 & 13.755 & 25.735 & \textbf{0.084} \\ % 1.503
    HyperTwinPeaks & 30 & 90.008 & 89.830 & 402.091 & 437.310 & \textbf{0.162} \\
    \bottomrule
  \end{tabular}
\end{table}
\section{Conclusion}
This work connects the adversarial vulnerability of score models with the geometry of the underlying manifold they capture. We show that minimizing the Dirichlet energy of learned score maps simultaneously boosts their robustness while revealing topological dimension. Leveraging this, we introduce a novel method to measure the topological dimension of manifolds captured by score models using adversarial attacks, thereby generating a useful interpretation of the data and model.
\newpage

\section*{Acknowledgements}

This material is based upon work supported by the U.S. Department of Energy, Office of Science, Office of Workforce Development for Teachers and Scientists, Office of Science Graduate Student Research (SCGSR) program. The SCGSR program is administered by the Oak Ridge Institute for Science and Education (ORISE) for the DOE. ORISE is managed by ORAU under contract number DE-SC0014664. All opinions expressed in this paper are the author’s and do not necessarily reflect the policies and views of DOE, ORAU, or ORISE.

This research is supported in part by U.S. Army Research funding W911NF2220025, U.S. Air Force Research Lab funding FA8750-21-1-1015. In addition, it is supported by the U.S. Department of Energy, through the Office of Advanced Scientific Computing Research’s “Data-Driven Decision Control for Complex Systems (DnC2S)” project.

This research employed resources of the Experimental Computing Laboratory (ExCL) at Oak Ridge National Laboratory.

\bibliography{neurips_2023}

\begin{thebibliography}{34}
\providecommand{\natexlab}[1]{#1}
\providecommand{\url}[1]{\texttt{#1}}
\expandafter\ifx\csname urlstyle\endcsname\relax
  \providecommand{\doi}[1]{doi: #1}\else
  \providecommand{\doi}{doi: \begingroup \urlstyle{rm}\Url}\fi

\bibitem[Campadelli et~al.(2015)Campadelli, Casiraghi, Ceruti, and
  Rozza]{campadelli2015intrinsic}
Paola Campadelli, Elena Casiraghi, Claudio Ceruti, and Alessandro Rozza.
\newblock Intrinsic dimension estimation: Relevant techniques and a benchmark
  framework.
\newblock \emph{Mathematical Problems in Engineering}, 2015:\penalty0 1--21,
  2015.

\bibitem[Pope et~al.(2021)Pope, Zhu, Abdelkader, Goldblum, and
  Goldstein]{pope2021intrinsic}
Phillip Pope, Chen Zhu, Ahmed Abdelkader, Micah Goldblum, and Tom Goldstein.
\newblock The intrinsic dimension of images and its impact on learning.
\newblock \emph{arXiv preprint arXiv:2104.08894}, 2021.

\bibitem[Rozza et~al.(2012)Rozza, Lombardi, Ceruti, Casiraghi, and
  Campadelli]{rozza2012novel}
Alessandro Rozza, Gabriele Lombardi, Claudio Ceruti, Elena Casiraghi, and Paola
  Campadelli.
\newblock Novel high intrinsic dimensionality estimators.
\newblock \emph{Machine learning}, 89:\penalty0 37--65, 2012.

\bibitem[Levina and Bickel(2004)]{levina2004maximum}
Elizaveta Levina and Peter Bickel.
\newblock Maximum likelihood estimation of intrinsic dimension.
\newblock \emph{Advances in neural information processing systems}, 17, 2004.

\bibitem[Camastra and Staiano(2016)]{camastra2016intrinsic}
Francesco Camastra and Antonino Staiano.
\newblock Intrinsic dimension estimation: Advances and open problems.
\newblock \emph{Information Sciences}, 328:\penalty0 26--41, 2016.

\bibitem[Song and Ermon(2019)]{song2019generative}
Yang Song and Stefano Ermon.
\newblock Generative modeling by estimating gradients of the data distribution.
\newblock \emph{Advances in neural information processing systems}, 32, 2019.

\bibitem[Song et~al.(2020)Song, Sohl-Dickstein, Kingma, Kumar, Ermon, and
  Poole]{song2020score}
Yang Song, Jascha Sohl-Dickstein, Diederik~P Kingma, Abhishek Kumar, Stefano
  Ermon, and Ben Poole.
\newblock Score-based generative modeling through stochastic differential
  equations.
\newblock \emph{arXiv preprint arXiv:2011.13456}, 2020.

\bibitem[Ho et~al.(2020)Ho, Jain, and Abbeel]{ho2020denoising}
Jonathan Ho, Ajay Jain, and Pieter Abbeel.
\newblock Denoising diffusion probabilistic models.
\newblock \emph{Advances in Neural Information Processing Systems},
  33:\penalty0 6840--6851, 2020.

\bibitem[Vignac et~al.(2022)Vignac, Krawczuk, Siraudin, Wang, Cevher, and
  Frossard]{vignac2022digress}
Clement Vignac, Igor Krawczuk, Antoine Siraudin, Bohan Wang, Volkan Cevher, and
  Pascal Frossard.
\newblock Digress: Discrete denoising diffusion for graph generation.
\newblock \emph{arXiv preprint arXiv:2209.14734}, 2022.

\bibitem[Vincent(2011)]{vincent2011connection}
Pascal Vincent.
\newblock A connection between score matching and denoising autoencoders.
\newblock \emph{Neural computation}, 23\penalty0 (7):\penalty0 1661--1674,
  2011.

\bibitem[Szegedy et~al.(2013)Szegedy, Zaremba, Sutskever, Bruna, Erhan,
  Goodfellow, and Fergus]{szegedy2013intriguing}
Christian Szegedy, Wojciech Zaremba, Ilya Sutskever, Joan Bruna, Dumitru Erhan,
  Ian Goodfellow, and Rob Fergus.
\newblock Intriguing properties of neural networks.
\newblock \emph{arXiv preprint arXiv:1312.6199}, 2013.

\bibitem[Yeats et~al.(2021)Yeats, Chen, and Li]{yeats2021improving}
Eric~C Yeats, Yiran Chen, and Hai Li.
\newblock Improving gradient regularization using complex-valued neural
  networks.
\newblock In \emph{International Conference on Machine Learning}, pages
  11953--11963. PMLR, 2021.

\bibitem[Lee et~al.(2023)Lee, Lee, Kim, Kim, and Uh]{lee2023diffusion}
Sangyun Lee, Gayoung Lee, Hyunsu Kim, Junho Kim, and Youngjung Uh.
\newblock Diffusion models with grouped latents for interpretable latent space.
\newblock In \emph{ICML 2023 Workshop on Structured Probabilistic Inference
  $\{$$\backslash$\&$\}$ Generative Modeling}, 2023.

\bibitem[Yeats et~al.(2022)Yeats, Liu, Womble, and Li]{yeats2022nashae}
Eric Yeats, Frank Liu, David Womble, and Hai Li.
\newblock Nashae: Disentangling representations through adversarial covariance
  minimization.
\newblock In \emph{Computer Vision--ECCV 2022: 17th European Conference, Tel
  Aviv, Israel, October 23--27, 2022, Proceedings, Part XXVII}, pages 36--51.
  Springer, 2022.

\bibitem[Pope et~al.(2020)Pope, Balaji, and Feizi]{pope2020adversarial}
Phillip Pope, Yogesh Balaji, and Soheil Feizi.
\newblock Adversarial robustness of flow-based generative models.
\newblock In \emph{International Conference on Artificial Intelligence and
  Statistics}, pages 3795--3805. PMLR, 2020.

\bibitem[Dherin et~al.(2022)Dherin, Munn, Rosca, and Barrett]{dherin2022neural}
Benoit Dherin, Michael Munn, Mihaela Rosca, and David Barrett.
\newblock Why neural networks find simple solutions: The many regularizers of
  geometric complexity.
\newblock \emph{Advances in Neural Information Processing Systems},
  35:\penalty0 2333--2349, 2022.

\bibitem[Madry et~al.(2017)Madry, Makelov, Schmidt, Tsipras, and
  Vladu]{madry2017towards}
Aleksander Madry, Aleksandar Makelov, Ludwig Schmidt, Dimitris Tsipras, and
  Adrian Vladu.
\newblock Towards deep learning models resistant to adversarial attacks.
\newblock \emph{arXiv preprint arXiv:1706.06083}, 2017.

\bibitem[Bac et~al.(2021)Bac, Mirkes, Gorban, Tyukin, and
  Zinovyev]{bac2021scikit}
Jonathan Bac, Evgeny~M Mirkes, Alexander~N Gorban, Ivan Tyukin, and Andrei
  Zinovyev.
\newblock Scikit-dimension: a python package for intrinsic dimension
  estimation.
\newblock \emph{Entropy}, 23\penalty0 (10):\penalty0 1368, 2021.

\bibitem[Koehler et~al.(2022)Koehler, Heckett, and
  Risteski]{koehler2022statistical}
Frederic Koehler, Alexander Heckett, and Andrej Risteski.
\newblock Statistical efficiency of score matching: The view from isoperimetry.
\newblock \emph{arXiv preprint arXiv:2210.00726}, 2022.

\bibitem[Krizhevsky et~al.(2009)Krizhevsky, Hinton,
  et~al.]{krizhevsky2009learning}
Alex Krizhevsky, Geoffrey Hinton, et~al.
\newblock Learning multiple layers of features from tiny images.
\newblock 2009.

\bibitem[Paszke et~al.(2019)Paszke, Gross, Massa, Lerer, Bradbury, Chanan,
  Killeen, Lin, Gimelshein, Antiga, et~al.]{paszke2019pytorch}
Adam Paszke, Sam Gross, Francisco Massa, Adam Lerer, James Bradbury, Gregory
  Chanan, Trevor Killeen, Zeming Lin, Natalia Gimelshein, Luca Antiga, et~al.
\newblock Pytorch: An imperative style, high-performance deep learning library.
\newblock \emph{Advances in neural information processing systems}, 32, 2019.

\bibitem[Chen(2018)]{torchdiffeq}
Ricky T.~Q. Chen.
\newblock torchdiffeq, 2018.
\newblock URL \url{https://github.com/rtqichen/torchdiffeq}.

\bibitem[Chen et~al.(2018)Chen, Rubanova, Bettencourt, and
  Duvenaud]{chen2018neuralode}
Ricky T.~Q. Chen, Yulia Rubanova, Jesse Bettencourt, and David Duvenaud.
\newblock Neural ordinary differential equations.
\newblock \emph{Advances in Neural Information Processing Systems}, 2018.

\bibitem[Carlini and Wagner(2017)]{carlini2017towards}
Nicholas Carlini and David Wagner.
\newblock Towards evaluating the robustness of neural networks.
\newblock In \emph{2017 ieee symposium on security and privacy (sp)}, pages
  39--57. Ieee, 2017.

\bibitem[Cohen et~al.(2019)Cohen, Rosenfeld, and Kolter]{cohen2019certified}
Jeremy Cohen, Elan Rosenfeld, and Zico Kolter.
\newblock Certified adversarial robustness via randomized smoothing.
\newblock In \emph{international conference on machine learning}, pages
  1310--1320. PMLR, 2019.

\bibitem[Khan and Storkey(2022)]{khan2022adversarial}
Asif Khan and Amos Storkey.
\newblock Adversarial robustness of $\beta-$ vae through the lens of local
  geometry.
\newblock \emph{arXiv preprint arXiv:2208.03923}, 2022.

\bibitem[Higgins et~al.(2017)Higgins, Matthey, Pal, Burgess, Glorot, Botvinick,
  Mohamed, and Lerchner]{higgins2017beta}
Irina Higgins, Loic Matthey, Arka Pal, Christopher Burgess, Xavier Glorot,
  Matthew Botvinick, Shakir Mohamed, and Alexander Lerchner.
\newblock beta-vae: Learning basic visual concepts with a constrained
  variational framework.
\newblock In \emph{International conference on learning representations}, 2017.

\bibitem[Yeats et~al.(2023)Yeats, Liu, and Li]{yeats2023disentangling}
Eric Yeats, Frank~Y Liu, and Hai Li.
\newblock Disentangling learning representations with density estimation.
\newblock In \emph{The Eleventh International Conference on Learning
  Representations}, 2023.

\bibitem[Nie et~al.(2022)Nie, Guo, Huang, Xiao, Vahdat, and
  Anandkumar]{nie2022diffusion}
Weili Nie, Brandon Guo, Yujia Huang, Chaowei Xiao, Arash Vahdat, and Anima
  Anandkumar.
\newblock Diffusion models for adversarial purification.
\newblock \emph{arXiv preprint arXiv:2205.07460}, 2022.

\bibitem[Wang et~al.(2022)Wang, Lyu, Lin, Dai, and Fu]{wang2022guided}
Jinyi Wang, Zhaoyang Lyu, Dahua Lin, Bo~Dai, and Hongfei Fu.
\newblock Guided diffusion model for adversarial purification.
\newblock \emph{arXiv preprint arXiv:2205.14969}, 2022.

\bibitem[Sohl-Dickstein et~al.(2015)Sohl-Dickstein, Weiss, Maheswaranathan, and
  Ganguli]{sohl2015deep}
Jascha Sohl-Dickstein, Eric Weiss, Niru Maheswaranathan, and Surya Ganguli.
\newblock Deep unsupervised learning using nonequilibrium thermodynamics.
\newblock In \emph{International Conference on Machine Learning}, pages
  2256--2265. PMLR, 2015.

\bibitem[Camastra and Vinciarelli(2002)]{camastra2002estimating}
Francesco Camastra and Alessandro Vinciarelli.
\newblock Estimating the intrinsic dimension of data with a fractal-based
  method.
\newblock \emph{IEEE Transactions on pattern analysis and machine
  intelligence}, 24\penalty0 (10):\penalty0 1404--1407, 2002.

\bibitem[Amsaleg et~al.(2015)Amsaleg, Chelly, Furon, Girard, Houle,
  Kawarabayashi, and Nett]{amsaleg2015estimating}
Laurent Amsaleg, Oussama Chelly, Teddy Furon, St{\'e}phane Girard, Michael~E
  Houle, Ken-ichi Kawarabayashi, and Michael Nett.
\newblock Estimating local intrinsic dimensionality.
\newblock In \emph{Proceedings of the 21th ACM SIGKDD International Conference
  on Knowledge Discovery and Data Mining}, pages 29--38, 2015.

\bibitem[Tempczyk et~al.(2022)Tempczyk, Michaluk, Garncarek, Spurek, Tabor, and
  Golinski]{tempczyk2022lidl}
Piotr Tempczyk, Rafa{\l} Michaluk, Lukasz Garncarek, Przemys{\l}aw Spurek,
  Jacek Tabor, and Adam Golinski.
\newblock Lidl: Local intrinsic dimension estimation using approximate
  likelihood.
\newblock In \emph{International Conference on Machine Learning}, pages
  21205--21231. PMLR, 2022.

\end{thebibliography}

\newpage
\appendix
\onecolumn

\section{Local Averaging Property of Score Map and Adversaries}\label{app:local_averaging}

In thermodynamics, the partition function of a physical system can be written in the form $e^{-u}$, where $u$ is interpreted as energy, which controls the physics of the system. Entropy is thought of as the loss of the ability to do work as the system becomes more disordered. First variation of entropy, indeed, can be seen as measuring the ability of one to shift material from areas of high potential to areas of low potential. That is, as the system mixes, entropy increases.

Log-likelihood adversaries, then, can be thought of as local thermodynamic exploits, with regions of high density immediately next to regions of low density \cite{pope2020adversarial}. In a physical system, this provides an opportunity to do work, while in a machine learning model, the ability to damage performance. By increasing entropy, one removes the opportunity to do work, hence removing the availability of adversaries.

There is a tradeoff, of course. Maximizing entropy will flatten the partition function out, and in practice this yields a model which will be far from maximizing log likelihood. Thought of in terms of score-based modeling, this means that globally minimizing the magnitude of the score in order to improve adversarial robustness will likely prevent the model from agreeing well with the true score.

In order to motivate the proposal to, instead, enforce a local averaging property for the score, as well as to provide a framework for eventually producing and proving rigorous theoretic results, we consider score-based modeling from the perspective of Langevin dynamics.

Suppose one begins with a sample drawn from the true distribution $e^{-u}$. Next, imagine one begins to perturb the data points according to a discrete approximation to a Brownian motion. That is, at every time step $t$, one moves each sample point by a perturbation drawn from a gaussian distribution of variance proportional to the size $\Delta t$ of the time steps. This will, over time, flatten the distribution out, maximizing entropy and removing adversaries, but also destroying the original partition function.

Consider, now, a single point in space, and a small codimension one disk centered at that point. Imagine that we count, as the Brownian motion continues, how frequently a data point moves on direction or the other across this disk. Now imagine that we did this simultaneously for every disk through that point. By measuring the net direction of the diffusion flow in this way, we might be able to introduce a drift term to counter-act this flow. That is to say, at each point in space, we may be able to add a bias to the gaussian used to generate the diffusion, which tends to push the density back to where it was flowing from, keeping everything stationary.

The Langevin equation is an SDE determined from a potential $u$:
\[
dX_t = -\nabla u\ dt + \sqrt{2} dW_t,
\]
where $dW_t$ is a Brownian motion. The unique stationary distribution for the Langevin equation is the partition function $e^{-u}$, and so we can intuitively interpret the score of a model with density $e^{-u}$ as the drift term envisioned before, needed as a correction to pure diffusion in order to keep the partition function stationary.

Intuitively, then, learning a score model is equivalent to counting how frequently sample points will cross different disks distributed at points in space as a Brownian motion evolves. We can interpret the difficulties encountered by score based models in this light: in large empty regions of of sample space, fewer disk-crossings will be counted, increasing variance and reducing the reliability of the model. Similarly, if the true distribution is concentrated near a positive codimension manifold, particles traveling from one point on the manifold, out into the off-manifold space, and then back onto the manifold, will appear to the manifold to have teleported, interfering with accurately counting disk-crossings in directions relevant to the actual score. As a side remark, this viewpoint also ties in with observations that isoperimetric inequalities \cite{koehler2022statistical}, which can intuitively be thought of as estimating the efficiency of searching out regions of space by sending out sample paths from a lattice of source points, can be used to give bounds on the statistical efficiency of score-based models.

Our proposal is, essentially, to speed up the mixing of the score model in regions of low density while giving higher weight to the naive score estimates near regions of high density.

Imagine a point in a region of low density. Fewer sample paths will pass near enough this point to interact with any disk through the point, so it will collect less data in providing an estimate of the score. This leads, of course, to higher variance in the score estimates nearby, providing the potential for adversaries.

Now consider, instead, a $\varepsilon$-ball centered at such a point. As $\varepsilon$ is taken larger, this ball encounters more and more sample paths, and hence has lower variance in estimating the net flux due to diffusion.  Recall, now, that the characteristic property of the score is that it provides the drift terms needed to counteract Brownian motion and keep the distribution stationary. As a larger number of sample paths proceed through this ball, their average flux will be determined by the average drift term within the ball: that is, the average score within the ball. Thus, by replacing the score at the center of the ball with the average score over the rest of the ball, one obtains a lower variance estimate of score which, in regions of low data, will still provide statistical characteristics close to the correct Langevin flow.

On the other hand, consider a point near high data where the true score map is not locally averaging. Imagine that we attempt to enforce, at such a point, an objective which forces the score to be locally averaging. Such a score will fail to maintain stationarity of the partition function, and, with a higher quantity of sample paths confirming this, it will do so to a higher degree of extremality. Tuning of hyperparameters which allows the dominance of local averaging in regions of low density, then, will fail exponentially fast (keeping in mind that the distributions are approximately gaussian) in dominating the score objective near high density data.

For any choice of hyperparameters, then, we expect that local averaging will become exponentially more dominant in low density regions, while the score objective will be exponentially more dominant in high density regions, allowing the optimization of the tradeoff of removing adversaries while maintaining performance of the score model.

\section{Link between Dirichlet Energy and Differential Entropy}\label{app:diff_ent}

The Dirichlet energy can be directly connected to differential entropy through the score of the Gaussian distribution. Consider the multivariate Gaussian density with log-probability:

\[
    \log p(x) = -\frac{k}{2}\log(2 \pi) - \frac{1}{2} \log\det (\Sigma) - \frac{1}{2}(x - \mu)^T\Sigma^{-1}(x - \mu).
\]

The \textit{score} is the gradient of the log-probability with respect to $x$:

\[
    \nabla_x\log p(x) = -\Sigma^{-1}(x-\mu). 
\]

This is a linear map with differential $-\Sigma^{-1}$. Reducing the Dirichlet energy of the map is equivalent to reducing the dominant singular value of $\Sigma^{-1}$ (i.e., increase the variance in that direction). Recall that the entropy of a multivariate Gaussian distribution is:

\[
    h(x) = \frac{k}{2}\log(2 \pi e) + \frac{1}{2}\log\det(\Sigma) = \frac{k}{2}\log(2 \pi e) - \frac{1}{2}\sum^k_{i=1} \log \sigma_i(\Sigma^{-1}),
\]

where $\sigma_i(\Sigma^{-1})$ is the $i$-th singular value of the inverse covariance matrix. Hence, reducing the Dirichlet energy of a map to a Gaussian distribution corresponds to increasing its entropy. Specifically, the entropy of the least entropic component is increased.

\section{Relation of Dirichlet Energy Regularization to the Learned Manifold}\label{app:additive_variance}

Given assumptions about the ambient structure of the data, we shall show that the level of Dirichlet energy regularization of the score model corresponds precisely to adding variance to the \textit{off-manifold} component of the learned density.

According to the manifold assumption, the data occupies a subset of the dimensions of the \textit{ambient} (observed) space, denoted locally as the \textit{manifold} (tangent) subspace $X_\mathcal{M}$. Furthermore, the \textit{off-manifold} distribution occupies a locally orthogonal subspace, denoted as the normal subspace $X_\perp$. We model the \textit{local} ambient distribution in local subspace coordinates as the product of the tangent subspace and normal subspace:

\[
    p(x) = p_{X_\mathcal{M},X_\perp}(x_\mathcal{M},x_\perp) = p_\mathcal{M}(x_\mathcal{M})p_{X_\perp}(x_\perp),
\]

where $x_\mathcal{M}$ and $x_\perp$ refer to the components of $x$ according to the local \textit{manifold} and \textit{off-manifold} orthogonal subspaces of $X$ at $x$. While the local \textit{off-manifold} distribution is typically unknown, a natural assumption is the isotropic Gaussian distribution, as it maximizes uncertainty of the \textit{off-manifold} distribution for a given variance. This is also a common assumption for image data which is the focus of this work.

Consider the explicit score matching objective (left side) and its tractable denoising counterpart (right side) proven by \cite{vincent2011connection}:

\begin{multline*}
    \mathbb{E}_t \left[ \lambda(t) \mathbb{E}_{x_0}\mathbb{E}_{x_t|x_0} \left[ \| s_\theta(x_t, t) - \nabla_{x_t} \log p_{t}(x_t) \|^2 \right] \right]\\
    = \mathbb{E}_t \left[ \lambda(t) \mathbb{E}_{x_0}\mathbb{E}_{x_t|x_0} \left[ \| s_\theta(x_t, t) - \nabla_{x_t} \log p_{0t}(x_t | x_0) \|^2 \right] \right] + C
\end{multline*}

We shall now investigate the derivatives of explicit score matching to show in the optimal setting that Dirichlet energy regularization corresponds exactly to adding variance to the learned off-manifold component when its distribution is isotropic Gaussian. 

Recall that Dirichlet energy regularization of the score map penalizes the maximal squared singular value(s) of the score differential. At a point $x$ on the manifold, a natural assumption is that the singular values of the true score map differential corresponding to the normal subspace are strictly larger than those of the tangent space (by the manifold assumption), and repeated (due to the isotropic Gaussian assumption). Hence, near the manifold, we can focus on the normal subspace when considering Dirichlet energy regularization, as only the maximal singular values of the score map differential contribute to the map's Dirichlet energy.

The Dirichlet energy regularization term is implemented via Jacobian power iteration with a Gaussian seed vector. Since power iteration projects a vector to the space spanned by the maximal eigenvectors, it can be thought of as regularizing each of the maximal eigenvalues of the Gram Jacobian (i.e., the squared singular values of Jacobian) with uniform probability. If there are $l$ repeated maximal singular values $\sigma^2_{\text{max},i}$ where $i=1,...,l$, we can consider their regularization at each iteration by random selection by a uniform categorical random variable $i \sim \text{Unif}(1,...,l)$. Taking expectation with respect to $i$ we have

\[
    \mathbb{E}_{i\sim\text{Unif}(1,...,l)} \left[ \sigma^2_{\text{max},i} \right] = (1/l)\Tr(\Sigma^2_\text{max}),
\]

where $\Tr(\Sigma^2_\text{max})$ is the trace of the diagonal matrix formed by the set of squared maximal singular values.

Therefore, if the score model locally captures an isotropic Gaussian distribution of dimension $n_\perp$, Dirichlet energy regularization of strength $n\gamma$ is equivalent to $(n/n_\perp) \gamma \Tr(\Sigma_\perp^2)$ in expectation, where $\Tr(\Sigma_\perp^2)$ is the trace of the set of maximal squared singular values of the score map Jacobian (at $x$).

For simplicity, let us analyze the regularized training criterion around a small neighborhood of $x$ in the normal subspace. If the neighborhood is sufficiently small, we may simply consider a linear approximation of the score model restricted to the normal subspace $s_\theta(x_\perp) \approx Ax_\perp + b$. The regularized training criterion then becomes: 

\[
    \mathbb{E}_{x_\perp\sim\text{Gauss}(0, \Sigma)}\left[ \| b + (A + \Sigma^{-1})x_\perp \|^2\right] + (n/n_\perp) \gamma \text{Tr}\{A'A\},
\]

with $n\gamma$ as the level of Dirichlet energy regularization of the score map and $\Sigma=\sigma^2 I$ is the covariance matrix of the centered Gaussian distribution in the normal subspace. First we shall show that the fixed point for $b$ is achieved when $b=0$. The first term of the training criterion can be rewritten as

\[
    \mathbb{E}_{x_\perp\sim\text{Gauss}(0, \Sigma)} \left[  \|b\|^2 + 2 b\cdot(A + \Sigma^{-1})x_\perp + \|(A + \Sigma^{-1})x_\perp\|^2  \right]
\]

Due to $\mathbb{E}[x_\perp]=0$, we can eliminate the middle term and take derivatives with respect to $b$

\[
    \frac{\partial}{\partial b}\mathbb{E}_{x_\perp\sim\text{Gauss}(0, \Sigma)} \left[  \|b\|^2 + \|(A + \Sigma^{-1})x_\perp\|^2  \right] = 2 b'
\]

Therefore $\|b\| \rightarrow 0$ during training. Omitting $b$ from further manipulations, we shall now characterize the training criterion with the SVD of $A = U_A \Sigma_A V_A'$:

%% \begin{multline*}
\begin{flalign*}
    &\mathbb{E}_{x_\perp\sim\text{Gauss}(0, \Sigma)}\left[ \| (A + \Sigma^{-1})x_\perp \|^2\right] + (n/n_\perp) \gamma \text{Tr}\{\Sigma_A^2\} \\
    &= \mathbb{E}_{x_\perp\sim\text{Gauss}(0, \Sigma)}\left[ \| A x_\perp + \Sigma^{-1}x_\perp \|^2\right] + (n/n_\perp) \gamma \text{Tr}\{\Sigma_A^2\} \\
    &= \mathbb{E}_{x_\perp\sim\text{Gauss}(0, \Sigma)}\left[ x_\perp'A'A x_\perp + x_\perp'(A' + A)\Sigma^{-1}x_\perp + x_\perp'\Sigma^{-1}\Sigma^{-1}x_\perp \right] + (n/n_\perp) \gamma \text{Tr}\{\Sigma_A^2\} 
\end{flalign*}
%% \end{multline*}

Considering each of the terms within the expectation separately and leveraging the fact that the distribution for $x_\perp$ is spherical, we have

\begin{flalign}\label{eqn:partial_trace}
    & \mathbb{E}_{x_\perp\sim\text{Gauss}(0, \Sigma)}\left[ x_\perp'A'A x_\perp + x_\perp'(A' + A)\Sigma^{-1}x_\perp + x_\perp'\Sigma^{-1}\Sigma^{-1}x_\perp \right] + (n/n_\perp) \gamma \text{Tr}\{\Sigma_A^2\} \notag \\
   & = \text{Tr}(\Sigma_A^2\Sigma + \Sigma^{-1} + (n/n_\perp) \gamma \Sigma_A^2) + \mathbb{E}_{x_\perp\sim\text{Gauss}(0, \Sigma)}\left[ x_\perp'(A' + A)\Sigma^{-1}x_\perp \right] \notag \\
    &= \text{Tr}(\Sigma_A^2(\sigma^2 + (n/n_\perp) \gamma) + \Sigma^{-1}) + \mathbb{E}_{x_\perp\sim\text{Gauss}(0, \Sigma)}\left[ x_\perp'(A' + A)\Sigma^{-1}x_\perp \right]
\end{flalign}

Let us focus on the rightmost term, within the expectation:

% \begin{multline*}
\begin{flalign*}
    & (A' + A)\Sigma^{-1} = x_\perp'(V_A \Sigma_A U_A' + U_A \Sigma_A V_A')\Sigma^{-1}x_\perp \\
    &= x_\perp'((V_A + U_A)\Sigma_A (V_A + U_A)' - U_A \Sigma_A U_A' - V_A \Sigma_A V_A') \Sigma^{-1}x_\perp
\end{flalign*}
% \end{multline*}

Let $K$ be the matrix $K = V_A + U_A$ with SVD formulation $K = U_K \Sigma_K V_K'$

\[
    (V_A + U_A)\Sigma_A (V_A + U_A)' = U_K \Sigma_K V_K' \Sigma_A V_K \Sigma_K U_K'
\]

We can again convert this expression to a trace by considering each term separately in the expectation:

\begin{equation}\label{eqn:final_term}
    \mathbb{E}_{x_\perp\sim\text{Gauss}(0, \Sigma)}\left[ x_\perp'(A' + A)\Sigma^{-1}x_\perp \right] = \text{Tr}(\Sigma_A(\Sigma_K^2 - 2I))
\end{equation}

Taking partial derivatives $\partial/\partial\Sigma_K$ and setting to 0, we have $\text{Tr}(\Sigma_A\Sigma_K) \rightarrow 0$ and therefore $\Sigma_A(\Sigma_K^2 - 2I) \rightarrow -2\Sigma_A$. We emphasize that if all singular values of $A$ are nonzero then $\Sigma_K^2 - 2I \rightarrow -2I$.

Putting (\ref{eqn:partial_trace}) and (\ref{eqn:final_term}) together, we derive:

\begin{flalign}
\label{eqn:full_trace}
    &\mathbb{E}_{x_\perp\sim\text{Gauss}(0, \Sigma)}\left[ \| (A + \Sigma^{-1})x_\perp \|^2\right] + (n/n_\perp) \gamma \text{Tr}\{\Sigma_A^2\} \\
    & = \text{Tr}(\Sigma_A^2(\sigma^2 + (n/n_\perp) \gamma) + \Sigma^{-1}) + \text{Tr}(\Sigma_A(\Sigma_K^2 - 2I))
\end{flalign}

Taking partial derivatives of (\ref{eqn:full_trace}) with respect to $\Sigma_A$ and setting the expression to 0, we have:

% \begin{multline}
\begin{flalign*}
    2\Sigma_A(\sigma^2 + (n/n_\perp) \gamma) + (\Sigma_K^2 - 2I) & = 0 \\
    \Sigma_A(\sigma^2 + (n/n_\perp) \gamma) & = I - \frac{1}{2}\Sigma_K^2 \\
    \Sigma_A & = \frac{I - \frac{1}{2}\Sigma_K^2}{\sigma^2 + (n/n_\perp) \gamma}
\end{flalign*}
% \end{multline}

Assuming all singular values of $A$ are nonzero, the optimal numerator converges to $I$. This convergence implies that $A$ is diagonal and reflecting (i.e., isotropic Gaussian) at the fixed point. We conclude that the score model approximates an off-manifold isotropic Gaussian distribution with variance $\sigma^2 + (n/n_\perp) \gamma$. \qed

\section{Empirical Reduction of Dirichlet Energy}\label{app:jac_pow_it}

We propose to reduce the Dirichlet energy of each score map in time via a power iteration on the Jacobian of the score network. Power iteration allows for computation of spectral norms without explicit knowledge of specific entries in the respective matrix. Output-side vector Jacobian products can be calculated efficiently using reverse-mode automatic differentiation. Furthermore, input-side vector Jacobian products can be estimated relatively efficiently via finite difference numerical approximations. Starting with a random vector, the vector Jacobian product can be iterated on between output and input with rescaling to estimate the maximal singular value. The vector returned by the power iteration is then applied as a scaled input perturbation to the score model, and the squared norm of the corresponding output perturbation is reduced during training. We use 5 iterations of Jacobian Power Iteration in all our experiments.

\begin{algorithm}
\caption{Jacobian Power Iteration}\label{alg:jac_pow_it}
\begin{algorithmic}
\Require $x \in \mathbb{R}^n,\ f: \mathbb{R}^n \rightarrow \mathbb{R}^m,\ b \in \mathbb{N}, h \in \mathbb{R}^+$ \Comment{$b$: iterations, $h$: size of finite differences}
\Ensure $\|v\|^2 \approx \sigma^2_\text{max}(df)$
\State $v \gets z \sim \text{Gaussian}(0,\,I)$ \Comment{Initial guess for $v \in \mathbb{R}^m$}
\For{$1,\,...,\,b$}
    \State $v \gets v / \|v\|$ \Comment{Rescale}
    \State $u \gets \nabla_x \left[ v'f(x) \right]$  \Comment{Automatic differentiation}
    \State $u \gets u / \|u\|$ \Comment{Rescale}
    \State $v \gets \left(f(x + u\frac{h}{2}) - f(x - u\frac{h}{2})\right) / h$ \Comment{Finite differences}
\EndFor
\State \textbf{return} $\|v\|^2$
\end{algorithmic}
\end{algorithm}

\section{Experiment Details} 

The continuous VP DDPM models of the CIFAR-10 \cite{krizhevsky2009learning} adversarial attack experiments were trained for 500000 iterations of batch size 128 and learning rate 2e-4 using the Adam optimizer. Code was modified from the Pytorch \cite{paszke2019pytorch} version of \citet{song2020score}. The hardware was 4 nVidia Tesla V100 32gb GPUs. The average BPD values were collected from the same subset of samples drawn from test set for each value of Dirichlet energy regularization. The number of samples was $2560$ for clean samples and $100$ for attacked samples (due to the expense of calculating adversarial examples for SBMs). 

Adversarial attacks are calculated by backpropagating gradients through the solution of the change of variables formula with the probability flow ODE \cite{song2020score}. We used the \textit{torchdiffeq} package \cite{torchdiffeq} with ``dopri5'' setting and atol and rtol set to 1e-3. This is an extremely expensive operation in terms of memory, constraining the batch sizes to 1 per GPU. For a PGD attack of 20 iterations, the adversarial example generation rate was roughly 6 per gpu-hour. There are likely more effective and efficient ways to generate adversarial examples for SBMs. We also tried the adjoint ODE method for the O(1) memory cost \cite{chen2018neuralode}, but the adjoint was too stiff and computationally expensive to be practically useful.

For topological dimension estimation experiments, we set the size of each dataset to $1000$ and normalized to have zero mean and unit variance. Our full DDPM and individual score map methods were trained with $20000$ iterations of batch size $64$ and learning rate 1e-3 with a cosine learning rate decay to 1e-5. The adversarial attack was PGD(10, $\sigma$), with $\gamma = 0.01$. We use 5 iterations of Jacobian power iteration in all our experiments. For full DDPM experiments with DE regularization, it is helpful to bias the variance schedule with a larger minimum variance (e.g., $\sigma_t^2=0.01$ when $t=0$).

\section{Related Work}

\paragraph{Adversarial Examples} \citet{szegedy2013intriguing} are considered the first to describe the phenomenon of \textit{adversarial examples}: imperceptible input perturbations that drastically change the output of a DNN. Since their conception, many methods to make more powerful adversarial examples \cite{carlini2017towards} and methods to defend DNNs against them \cite{madry2017towards, yeats2021improving} have been developed. Adversarial training is a brute-force method which trains neural networks on self-crafted adversarial examples, which typically requires an inner attack optimization loop \cite{madry2017towards}. Although it is computationally demanding, adversarial training has been the most successful in practice. Gradient regularization encourages DNNs to be more resistant to adversarial examples by penalizing the norm of input gradients with respect to the loss \cite{yeats2021improving}. Randomized smoothing seeks to train a neural network on noisy inputs, and noisy inputs are again presented during inference. The output of the neural network is a volume-based classification represented by the summary of the outputs of many noisy samples \cite{cohen2019certified}.

While adversarial attacks have been studied extensively for discriminative models, exploration of adversarial attacks against generative models has been comparatively limited. A few works study the robustness of VAE reconstructions and encodings. \citet{khan2022adversarial} draw an interesting connection between the robustness of VAE models and the disentanglement of their latent space \cite{higgins2017beta,yeats2023disentangling}. \citet{pope2020adversarial} study adversarial attacks normalizing flow models, suggesting a hybrid adversarial training approach to boost robustness. A few works have employed the denoising capabilities of diffusion models as a \textit{defense tactic} \cite{nie2022diffusion, wang2022guided}, but their own vulnerability to adversarial samples (in terms of likelihood) is largely unexplored.

\paragraph{Diffusion Models} \citet{sohl2015deep} spurred modern interest in diffusion models - deep neural networks which learn a generative distribution by gradually uncorrupting noise from a prior distribution into samples of an arbitrarily complex distribution. The learning process consists of corrupting samples from a dataset and fitting a corresponding denoising transformation. \citet{ho2020denoising} presented adjustments to the denoising method to significantly improve the generation quality of images, achieving state-of-the-art in FID score on the CIFAR-10 dataset \cite{krizhevsky2009learning}. \citet{song2020score} extend the framework of diffusion models to continuous stochastic dynamics, invertible deterministic parameterization, and exact likelihood computation. Their continuous models achieve state-of-the-art in likelihood scores on various computer vision datasets. We focus on the \textit{score-based models} of \cite{song2020score} for their continuous formulation and exact likelihood computation capabilities.

\paragraph{Topological Dimension Estimation} The goal of topological dimension estimation (or local intrinsic dimension estimation) is to find the minimal number of variables needed to locally represent data without (much) information loss \cite{camastra2016intrinsic}. This has obvious implications in data compression, but also machine learning \cite{campadelli2015intrinsic}. Existing local estimators rely on fitting the fractal dimension \cite{camastra2002estimating}, using extreme value theory \cite{rozza2012novel}, or fitting statistical models (e.g., Poisson) \cite{levina2004maximum}. With extreme value theory, empirical distributions of the distances of points neighboring a given point are created, and the rate at which that density grows with a cutoff distance is used to estimate dimension \cite{amsaleg2015estimating}. For the Poisson method introduced by \citet{levina2004maximum}, the maximum likelihood estimator is developed for $k$-NN samples as the inverse of the average log-ratio of distances of $k-1$ neighbors with the $k$-th neighbor. \citet{tempczyk2022lidl} introduce an interesting method to estimate topological dimension using normalizing flows. Their core idea is that fitting flows to the data perturbed by Gaussian noise at different scales should result in differences in log density that are proportional to topological dimension. 

\section{Limitations}

The proposed topological dimension estimation method was competitive with existing statistical methods on the low-dimensional manifolds and most accurate on the noisy and higher-dimensional manifolds. However, the accuracy of the proposed method is contingent on two conditions being met: 1) the score model must fit the data very well and 2) the score model must reach the fixed point outlined in Theorem (\ref{thm:additive_variance}). If the second condition isn't met, the topological dimension estimates are not valid. If the first condition is not met, the topological dimension estimates may reflect those of the manifold captured by the model, but not necessarily that of the source data.

\end{document}